\documentclass[conference]{IEEEtran}
\IEEEoverridecommandlockouts
\usepackage{cite}
\usepackage{amsmath,amssymb,amsfonts}
\usepackage{algorithmic}
\usepackage{graphicx}
\usepackage{textcomp}
\usepackage{xcolor}
\usepackage{multirow}
\def\BibTeX{{\rm B\kern-.05em{\sc i\kern-.025em b}\kern-.08em
    T\kern-.1667em\lower.7ex\hbox{E}\kern-.125emX}}
\begin{document}

\title{Unlabeled Data Guided Semi-supervised \\ Histopathology Image Segmentation\\
\thanks{978-1-7281-6215-7/20/\$31.00 ©2020 IEEE.}
}

\author{\IEEEauthorblockN{Hongxiao Wang}
\IEEEauthorblockA{
% \textit{Dept. of Computer Science and Engineering} \\
\textit{University of Notre Dame}\\
Notre Dame, IN 46556, USA\\
hwang21@nd.edu \\ 
}
\IEEEauthorblockN{Lin Yang}
\IEEEauthorblockA{
% \textit{Dept. of Computer Science and Engineering} \\
\textit{University of Notre Dame}\\
Notre Dame, IN 46556, USA\\
lyang5@nd.edu}

\and
\IEEEauthorblockN{Hao Zheng}
\IEEEauthorblockA{
% \textit{Dept. of Computer Science and Engineering} \\
\textit{University of Notre Dame}\\
\ \ \ Notre Dame, IN 46556, USA\ \ \\
hzheng3@nd.edu}
\IEEEauthorblockN{Yizhe Zhang}
\IEEEauthorblockA{
% \textit{Dept. of Computer Science and Engineering} \\
\textit{University of Notre Dame}\\
\ \ \ Notre Dame, IN 46556, USA\\
yzhang29@nd.edu}
\and
\IEEEauthorblockN{Jianxu Chen}
\IEEEauthorblockA{
% \textit{dept. name of organization (of Aff.)} \\
\textit{ \ \ \ Allen Institute for Cell Science \ \ \ }\\
Seattle, WA 98103, USA\\
jianxuc@alleninstitute.org}
% \\\and
% \and
% \and
\IEEEauthorblockN{Danny Z. Chen}
\IEEEauthorblockA{
% \textit{Dept. of Computer Science and Engineering} \\
\textit{University of Notre Dame}\\
Notre Dame, IN 46556, USA\\
dchen@nd.edu}
}
\vspace{0.15in}
\maketitle

\begin{abstract}
Automatic histopathology image segmentation is crucial to disease analysis. Limited available labeled data hinders the generalizability of trained models under the fully supervised setting. Semi-supervised learning (SSL) based on generative methods has been proven to be effective in utilizing diverse image characteristics. However, it has not been well explored what kinds of generated images would be more useful for model training and how to use such images.
In this paper, we propose a new data guided generative method for histopathology image segmentation by leveraging the unlabeled data distributions. First, we design an image generation module. Image content and style are disentangled and embedded in a clustering-friendly space to utilize their distributions. New images are synthesized by sampling and cross-combining contents and styles. Second, we devise an effective data selection policy for judiciously sampling the generated images: (1) to make the generated training set better cover the dataset, the clusters that are underrepresented in the original training set are covered more; (2) to make the training process more effective, we identify and oversample the images of ``hard cases'' in the data for which annotated training data may be scarce.
Our method is evaluated on glands and nuclei datasets. We show that under both the inductive and transductive settings, our SSL method consistently boosts the performance of common segmentation models and attains state-of-the-art results.

\end{abstract}


\section{Introduction}
Deep learning methods have achieved unprecedented high performance in segmenting histopathology images \cite{yang2017suggestive,graham2019mild}. Still, the generalizability of such methods is hindered by limited available annotated training data, because medical experts are commonly required for annotation and there are high variations of image characteristics due to scanner effects, different staining protocols, patients, and disease states.
With limited annotation, deep learning model training often covers only a limited fraction of the histopathology data space. There could exist considerable discrepancy (in appearance) between the labeled and unlabeled sets. Thus, the trained segmentation models are at risk of over-fitting and do not generalize well to unseen data.
% possible data acquisition conditions change (e.g.,

An array of methods based on semi-supervised learning (SSL) has been proposed to make the most of limited training data and improve the model generalizability. The assumption is that unlabeled images are commonly from the original data distribution and contain useful information. In practice, there is often a large amount of unlabeled data available which are free to use. 
Some powerful SSL methods used the feature distribution of unlabeled images to reduce the need for labeling. For example, images are projected to low-dimensional feature space and pseudo-labels are assigned to unlabeld images based on clustering features \cite{iscen2019label,shi2018transductive}. In \cite{li2020semi,he2018decision}, images were intentionally perturbed to explore the decision boundary for adversarial training. While such SSL methods are commonly used for classification tasks, applying them to segmentation tasks is not a straightforward process because it is hard to define and utilize the distribution/clusters of unlabeled images due to the high-dimensional feature space of images. 
Dominant SSL methods for segmentation include using auxiliary loss \cite{Weakly}, consistency learning \cite{french2019semi}, and pseudo-labeling \cite{wei2018revisiting}. For example, auxiliary loss was applied to encourage the model to produce output of plausible shapes \cite{Weakly}; an ensemble method \cite{zheng2019new} was proposed to train a meta-learner with generated pseudo-labels; some methods \cite{french2019semi,cct,mixmatch} aimed to make the models give consistent predictions on random perturbations (e.g., color jitting, rotation). For histopathology images, the aforementioned methods have a main drawback: \emph{Image characteristics different from the labeled samples cannot be 
effectively and efficiently utilized in the training process}. 

To deal with diverse characteristic variations of unlabeled data, 
an intuitive and effective way is to adapt style-transfer based data generation methods for SSL  \cite{bentaieb2017adversarial,shaban2018staingan,ma2019neural}. Specifically, these methods generated new $(image, mask)$ pairs by transferring styles extracted from unlabeled images to labeled images. However, there are two unexplored issues in the previous work. 
First, neither image-based \cite{photowct,gatys} nor domain-based \cite{munit,CycleGAN2017} style transfer methods could concurrently provide a lightweight style representation and transfer styles between images in one (the same) domain, which is important for exploiting intrinsic data diversity and distributions of one dataset. 
Second, the associated augmentation policies of the known data generation methods were not carefully designed. For example, the methods \cite{tamura2019augmented,xue2019selective} filtered the generated images with higher feature similarity to the training images; in \cite{cubuk2018autoaugment,ho2019population}, complex training procedures were leveraged to search for an optimal combination of very basic image augmentation operations (e.g., rotation, flip, color jitting). 
These policies did not consider characteristics of unlabeled data and the discrepancy between labeled and unlabeled data, which can provide helpful information for data generation. 

In this paper, we propose a unified framework to exploit image characteristics and effectively employ it to guide data generation. 
We contemplate two main challenges: (1) image distribution is hard to observe and explicitly define; (2) an appropriate data generation policy needs to be designed for sampling from the augmented dataset to assist segmentation model training efficiently.

\begin{figure}[!t]
    \centering
    \includegraphics[width=\columnwidth,height = 2.8cm]{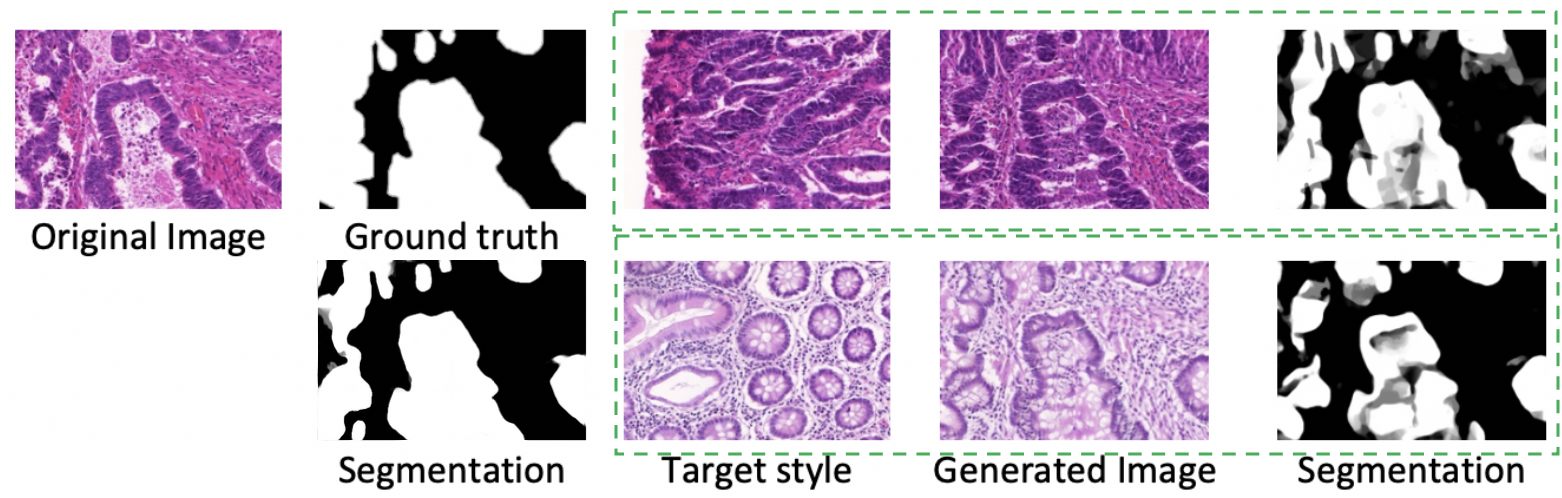}
    \caption{Examples of style transferred images and segmentation results of DCN \cite{dcn}. Different styles affect the segmentation performance of DCN.}
    \label{fig:1}
\end{figure}

Suppose our dataset consists of labeled and unlabeled sets. To address the first challenge, we exploit the relation between segmentation and image characteristics. We develop an Image Generation Module to learn and disentangle image representations and define characteristics distributions. 
Specifically, we consider two complementary key image characteristics in histopathology images: style and content. 
Style variation is caused by technical issues, including variations of staining protocols and scanner effects. Although it may not affect human judgment a lot, it can lead to reduced deep learning segmentation model performance (e.g., see Fig.~\ref{fig:1}). 
Content variation is directly related to specific segmentation tasks, which attributes to object shapes and distribution patterns related to disease types. 
Our module disentangles each histopathology image into style and content representations, whose distributions can be easily exploited by clustering. 
To make the features clustering-friendly, we embed style and content into low-dimensional spaces, for which an interpolation property is enforced to explore image similarity distances. 
We then generate new images by combining judiciously-selected pairs of style and content. 
% During model training, we extract content only from the labeled set, such that the pairs $(generated\ image, label)$ can be used.

For the second challenge, we develop a new data generation policy to handle underrepresented and ``hard-case'' images (which usually impair model generalizability the most).
To remedy the unbalanced data distributions in the labeled set, we propose a \emph{distribution matching strategy} to match the statistics of the labeled set with those of the unlabeled set across clusters of images by adding generated images to the underrepresented clusters. 
Moreover, ``hard-cases" are highly valuable for training networks but are often only a small fraction of the whole dataset and not sufficiently represented in the labeled data. For this, we propose a \emph{hard-case covering strategy} to identify hard cases via output variations in terms of different style changes and oversample them by referring to the unlabeled data. 
To make the generated images meaningful to segmentation, we only transfer styles between images that are sampled from the same content cluster.

We conduct extensive experiments on two public histopathology datasets of nuclei~\cite{monuseg} and glands~\cite{glas}, which show that our new SSL method achieves large segmentation improvement using common segmentation models through data generation. For both inductive and transductive settings, we attain state-of-the-art performance.

\section{Method}\label{sec:method}
\begin{figure*}[t]
    \centering
    \includegraphics[width=18cm,height = 6.2cm]{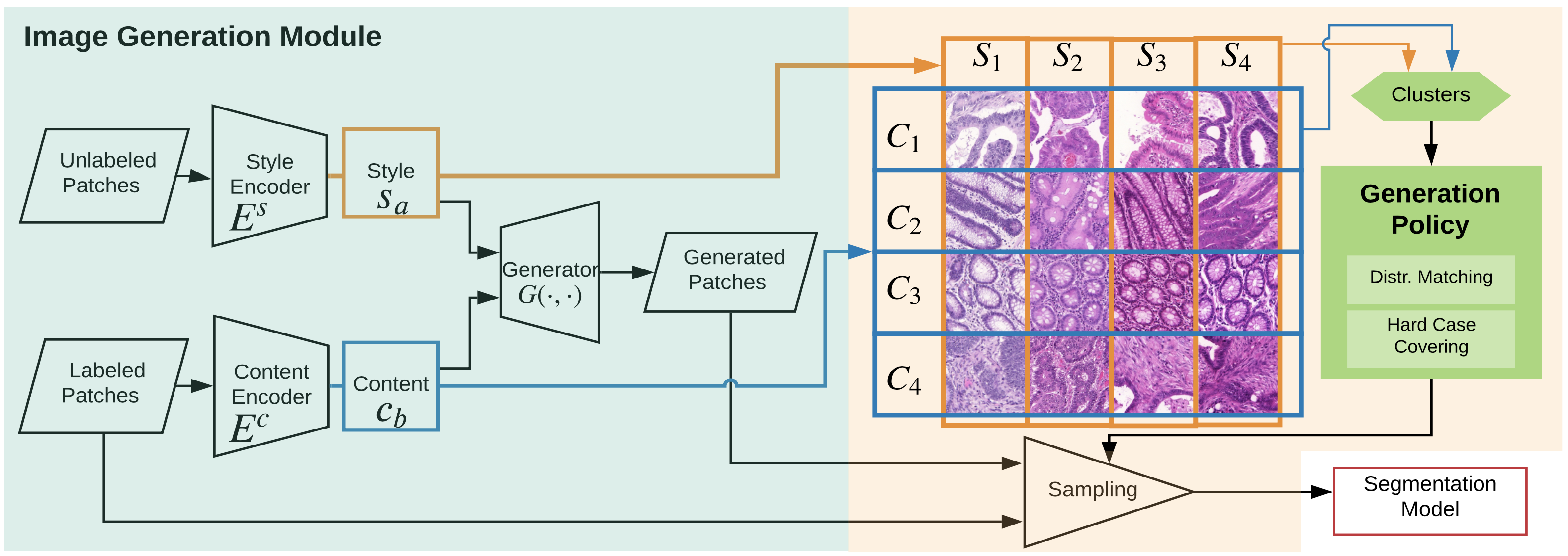}
    \caption{The pipeline of our proposed method.}
    \label{fig:framework1}
\end{figure*}

Fig.~\ref{fig:framework1} gives an overview of our proposed SSL framework, which consists of two parts. First, we introduce our image generation module with an interpolation property enforced in latent space to exploit image characteristics distributions. Second, using the image generation module, we propose a new strategy to sample from the generated data to bridge the distribution discrepancy between the labeled and unlabeled data, and a strategy to identify and handle the hard cases for segmentation based on a devised uncertainty metric.
% \vspace{0.15in}
% based on segmentation model variation to the input style changes

\subsection{Image Generation Module}
MUNIT~\cite{munit} is a state-of-the-art approach for cross-domain transformation, which explicitly disentangles image representation into content and style and provides a certain degree of explainability of the extracted latent image representation. 
It assumes that the source domain and target domain have \emph{distinguishable} style spaces. But, in our problem, this is not the case (i.e., our source and target domains significantly overlap), and MUNIT cannot yield diversified images by transferring style from one image to another since both images are from the same dataset.
Image-based methods can transfer styles between two arbitrary images; but, their representations of style and content are both in high-dimensional latent space (e.g., feature maps \cite{gatys} or matrices \cite{wct,photowct}), and thus it is hard to explore the image characteristics distributions. We will show how to utilize domain-based and image-based methods to (1) generate style-diversified images for model training and (2) make the image characteristics concisely represented so that their distributions can be effectively explored for segmentation.

% \begin{figure}[!t]
%     \centering
%     \includegraphics[width=9cm,height = 8.7cm]{supp1_.pdf}
%     \caption{Illustrating the losses for training our Image Generation Module.}
%     \label{fig:framework2}
% \end{figure}

We extract features with an Image Generation Module.
To capture the local contents, we consider image characteristics distributions on image patches uniformly cropped.
% I modify all subscript to a,b when it refers to patches, to i,j,k,l when it refers to clusters.
Given a set of $M$ image patches, $\mathcal{X} = \{x_a\}_{a=1}^M$, from our dataset, each patch $x_a$ can be encoded into a style vector $s_a=E^s(x_a)$ by a style encoder $E^s$ and a content vector $c_a=E^c(x_a)$ by a content encoder $E^c$. All the $s_a$'s and $c_a$'s form a style space $S=\{s_a\}_{a=1}^M$ and a content space $C=\{c_a\}_{a=1}^M$, respectively. A new image patch can be synthesized by a generator $G(c_a,s_b)$ combining any content vector $c_a$ and style vector $s_b$.
The total loss consists of our style matching loss, GAN loss, and reconstruction loss \cite{munit}, defined as:
% Given a set of image patches, $\mathcal{X} = \{x_i\}_{i=1}^M$, from our dataset, each patch $x_i$ can be encoded into a style vector $s_i=E^s(x_i)$ by a style encoder $E^s$ and a content vector $c_i=E^c(x_i)$ by a content encoder $E^c$. All the $s_i$'s and $c_i$'s form a style space $S=\{s_i\}_{i=1}^M$ and a content space $C=\{c_i\}_{i=1}^M$, respectively. A new image patch can be synthesized by a generator $G(c_i,s_j)$ combining any content vector $c_i$ and style vector $s_j$. The total loss consists of our style matching loss, GAN loss, and reconstruction loss \cite{munit} which is defined as:

\begin{equation}
     L = w_1 L_{style} + w_2 L_{GAN} + w_3 L_{recon}
\label{eq:totalLoss}
\end{equation}
where the $w_i$'s are hyper-parameters for controlling the weight of each term. We design the style matching loss for achieving two goals: (1) preventing the training collapse (which would cause all information being encoded into content and the generator ignoring the style); (2) making the point distribution  $\{s_a\}_{a=1}^M$ in the low-dimensional latent space $S$ reflect the image style distribution. 
The core of the style matching loss is an interpolation relation, which associates the point-wise distance in the embedded space with the similarity between image patches. 

To attain the first goal, when transferring a patch $x_a$ to the style $s_b=E^s(x_b)$ of a patch $x_b$, we encourage the generated patch $x_{g1} = G(E^c(x_a),s_b)$ to have the target style $s_b$, by minimizing a style similarity metric \cite{gatys} between $x_{g1}$ and $x_b$: 
% \begin{equation}
%   L_{s}(x_{g1},x_b) = \sum^L_{\ell=1}\frac{\alpha_{\ell}}{2N^2_{\ell}}\sum_{ij}(J_{\ell}[x_{g1}]-J_{\ell}[x_{b}])_{ij}^2
% \label{eq:gatys}
% \end{equation}
\begin{equation}
  L_{s}(x_{g1},x_b) = \sum^L_{\ell=1}\frac{\alpha_{\ell}}{2N^2_{\ell}}||J_{\ell}[x_{g1}]-J_{\ell}[x_{b}]||^2
\label{eq:gatys}
\end{equation}
where $J[x]$ is a Gram matrix calculated on the vectorized VGG features of layer $\ell$, $\alpha_{\ell}$ is the weight of layer ${\ell}$, and $N_{\ell}$ is the number of filters in layer ${\ell}$.

For the second goal, we encourage that two patches with a higher similarity attain a smaller distance in $S$. We train the model by enforcing the interpolation property: For a linear interpolation in the latent space, $s_{g2}=(1-\lambda) E^s(x_a)  + \lambda E^s(x_b)$, where $\lambda \in [0,1]$, the style of the generated patch $x_{g2}=G(E^c(x_a),s_{g2})$ smoothly transfers to $s_b$ as the distance between $s_{g2}$ and $s_{b}$ gets closer. This property can be learned by optimizing the following style matching loss: % is defined as:
\begin{equation}
     L_{style} =   \mathbb{E}_{x_a,x_b,\lambda}||(1-\lambda) L_{s}(x_{g2},x_a) - \lambda L_s(x_{g2},x_b)||
     \label{eq:loss2}
\end{equation}
% we interpolate $s_{g2}=(1-\lambda) E^s(x_a)  + \lambda E^s(x_b))$ in style embedding space.
where $\lambda$ is uniformly selected from $[0,1]$ in training. 
By the interpolation property, those patches with similar styles tend to have closer style vectors, yet dissimilar patches tend to have style vectors far apart from one another. Hence, the distribution of style vectors in the style space is encouraged to reflect the patch style distribution in the given dataset. Only Eq.~(\ref{eq:loss2}) is practically used in our training, as Eq.~(\ref{eq:gatys}) is a special case of Eq.~(\ref{eq:loss2}) with $\lambda=1$.

% I modify all subscripts to a,b when it refers to patches (upper bound is M), to i,j,k,l when it refers to clusters (upper bounds are m,n,m,n respectively).
After obtaining the extracted content set $C=\{c_a\}_{a=1}^M$ and style set $S=\{s_a\}_{a=1}^M$ from all the patches, we conduct agglomerative clustering to attain content clusters $\{C_i\}_{i=1}^m$ of $C$ and style clusters $\{S_j\}_{j=1}^n$ of $S$, where $m$ and $n$ are the numbers of such clusters, respectively. A patch space $H$ is defined by the Cartesian product of $\{C_i\}_{i=1}^m$ and $\{S_j\}_{j=1}^n$: $H = \{C_i\}_{i=1}^m \times \{S_j\}_{j=1}^n = \{H_{ij} = \{(c,s) \ | \ c \in C_i, s \in S_j\}\}$. Each $H_{ij}$ corresponds to possible patches whose content vectors belong to content cluster $C_i$ and style vectors belong to style cluster $S_j$.
We explore the \emph{statistics} of the patch space $H$ based on the information of each $H_{ij}$, whose numbers of labeled patches and unlabeled patches are denoted by $N(H_{ij}^{label})$ and $N(H_{ij}^{unlabel})$, respectively. 
% After obtaining the extracted content set $C=\{c_i\}_{i=1}^M$ and style set $S=\{s_i\}_{i=1}^M$ from all the patches, we conduct agglomerative clustering to attain content clusters $\{C_i\}_{i=1}^m$ of $C$ and style clusters $\{S_j\}_{j=1}^n$ of $S$, where $m$ and $n$ are the numbers of such clusters, respectively. A patch space $H$ is defined by the Cartesian product of $\{C_i\}_{i=1}^m$ and $\{S_j\}_{j=1}^n$: $H = \{C_i\}_{i=1}^m \times \{S_j\}_{j=1}^n = \{H_{ij} = \{(c,s) \ | \ c \in C_i, s \in S_j\}\}$. Each $H_{ij}$ corresponds to possible patches whose content vectors belong to content cluster $C_i$ and style vectors belong to style cluster $S_j$.
%We explore the \emph{statistics} of the patch space $H$ based on the information of each $H_{ij}$, whose numbers of labeled patches and unlabeled patches are denoted by $N(H_{ij}^{label})$ and $N(H_{ij}^{unlabel})$, respectively. 

% \subsubsection{Reconstruction Loss}
The same reconstruction loss for images and the latent space as in MUNIT \cite{munit} is applied to the generated images. The content and style should be consistent after decoding and encoding. The latent reconstruction loss is computed as:
\begin{align}
        L_{recon} & = L^x_{recon} + L^c_{recon} + L^s_{recon} \\
    {\rm with:} \
     L^x_{recon} & = \mathbb{E}_{x} ||x - G(E^c(x),E^s(x))||_{1}  \\
     L^c_{recon} & = \mathbb{E}_{c,s} ||c - E^c(G(c,s))||_{1}  \\
     L^s_{recon} & = \mathbb{E}_{c,s} ||s - E^s(G(c,s))||_{1}   
\end{align}
where $L^x_{recon}$, $L^c_{recon}$, and $L^s_{recon}$ are the image, content, and style reconstruction losses, respectively. 

We seek to attain appearance realism of the generated images by training the discriminator using samples generated with interpolated styles. The generated image distribution is imposed to be the same as the original image distribution. The realism loss function is:
\begin{equation}
L_{GAN}=\min_{E,G}\max_{D}\mathbb{E}_{x_a,s}[\log(1 - D(G(c_a,s))) + \log(D(x_a))]
\end{equation}
where $x_a$ follows the original image distribution ($x_a\sim p(x)$), $s=(1-\lambda) * s_a + \lambda * s_b$ with $s_a,s_b \sim p(s)$, and $c_a$ is encoded from $x_a$ with $c_a=E^c(x_a)$.

\subsection{Generation Policy}
Using extracted $c_i\in C$ and $s_j\in S$, we can generate a set of patches that may potentially help segmentation models in a basic \emph{random generation} setting (by uniform sampling). However, by doing so, (1) biomedically invalid patches may be generated without considering content similarity between the source images and target images, and (2) underrepresented and rare hard cases are overwhelmed by other ``common" image characteristics. 
In this section, we first discuss a basic random generation strategy with content matching, and then further improve the effectiveness by proposing our distribution matching policy and hard case covering policy.

\noindent
{\bf Random Generation with Content Matching.}\label{ra}
In common practice, one may generate data by transferring a labeled image to any other styles with the original segmentation label preserved. This generates an \emph{augmented set} (denoted as $S_{gen}$) from the original labeled set ($S_{label}$). 
However, invalid images can be generated when a target style is not biomedically valid with respect to the original content (e.g., an image of irregular-shaped cancerous glands with benign tubular texture). To handle this, we propose a \emph{content matching strategy} that exchanges styles only between patches from the same content cluster, as follows:
\begin{equation}
\begin{aligned}
     S_{gen}&  (  S_{label}, S_{label}\cup S_{unlabel}  ) =  \{   G ( E^c (x_a), E^s(x_b))  \\
    &  |\phantom{x} \forall x_a \in S_{label}, \forall x_b \in S_{label}\cup S_{unlabel}, \\    & E^c (x_a) \ \& \ E^c (x_b)\ 
 \text{are in the same content cluster}\} 
\end{aligned}
\end{equation}
 
Still, $|S_{gen}| \gg |S_{label}|$, and we need to strike a balance between the effectiveness of the augmented set and avoiding excessively perturbing the original labeled (training) data. $S_{gen}$ is used with a probability $R_{a}$ (a probability $1-R_{a}$ for $S_{label}$).

\noindent
{\bf Policy 1: Generation with Distribution Matching.}
To improve segmentation performance, we seek to remedy the observed statistics discrepancy between the labeled and unlabeled sets (i.e., $H_{ij}^{label}$ and $H_{ij}^{unlabel}$). %
%, such that underrepresented clusters can be better covered. 
% Fig.~\ref{fig:3}(A) shows an example of statistics discrepancy in the GlaS dataset~\cite{glas}, where the original training set is used as labeled data and the original test set is taken as unlabeled data; 
In preliminary study, we found that the numbers of patches in different $H_{ij}$'s are highly unbalanced and show significant statistic differences between the labeled and unlabeled sets. Randomly sampling patches could bury rare image characteristics into other image characteristics, causing the segmentation models to be trained inadequately.

Thus, we propose to sample patches based on the statistics of the unlabeled data. Especially, we focus on the underrepresented clusters with image characteristics that appear frequently in the unlabeled set but rarely in the labeled set. Each $H_{ij}$ is selected with a probability $N(H^{unlabel}_{ij})/{\Sigma_{k,l}N(H^{unlabel}_{kl})}$. Then we uniformly select samples from $S_{gen}\cap H_{ij}$ (or $S_{label}\cap H_{ij}$) with probability $R_a$ (or $1-R_a$). In this way, the underrepresented clusters benefit the most as they get higher chances to be included in the augmented training data.

\noindent
{\bf Policy 2: Generation with Hard Case Covering.}
To improve segmentation, we further identify and handle the ``hard cases". In our preliminary experiments, we found that the segmentation performances across different clusters can be different. For example, the performance is not good enough for irregular-shaped or highly stained abnormal tissues for all the segmentation models that we considered. 

To address this issue, we propose to identify these hard cases automatically. Inspired by bootstraping-based methods \cite{johnson2001introduction,yang2017suggestive}, we quantify the uncertainty of a segmentation model by its prediction variance when the input is transferred to different target styles. For $n$ style clusters $\{S_j\}^n_{j=1}$, we select one representative target style $Rep(S_l)$ from each $S_l$ with the minimum sum of distances to all the other styles $s_k\in S_l$. For each $H_{ij}$, its uncertainty $U_{ij}$ for a segmentation model $Seg$ is calculated as: 
\begin{equation}
\begin{aligned}
U_{ij} = & \frac{1}{N(H^{unlabel}_{ij})}  \sum_{x_a\in H^{unlabel}_{ij}, \ l \in\{1,2,\ldots,n\}} \\
 &    Variance_{(l)}(Seg(G(E^c(x_a),Rep(S_l))))
\end{aligned}
\end{equation}

% As shown in Fig.~\ref{fig:3}(B), the uncertainty has strong correlation with the pixel-wise F1 scores of segmentation for the unlabeled data in the GlaS dataset \cite{glas} (e.g., its coefficient of determination is $R^2 =0.57$).
Selecting uncertain clusters more frequently will help the segmentation model reduce potential prediction errors. Thus, we upweight the probability of sampling uncertain clusters. Each $H_{ij}$ is selected with a probability $U_{ij}/{\Sigma_{k,l}U_{kl}}$.
This policy has two advantages comparing to \cite{yang2017suggestive}: (1) our method does not require training separate versions of the segmentation model; (2) 
% comparing to \cite{yang2017suggestive}, 
our uncertainty value can better assess the segmentation quality, as the experiments show that our coefficient $R^2$ is 15\% higher.

Overall, these two policies 
%we propose 
are complementary to each other and can be combined with \emph{Mixed policy}: sample patches with an average probability of distribution matching and hard case covering (i.e., the probability is $N(H^{unlabel}_{ij})/(2 \Sigma_{k,l}N(H^{unlabel}_{kl}))+U_{ij}/(2 \Sigma_{k,l}U_{kl})$), which comprehensively covers the unlabeled data especially for the hard cases.
% (2) \emph{Ensemble policy}: average-ensemble the results of the two models, each trained using one of the two policies. 

\section{Experiments}\label{sec:exp}
\subsection{Datasets and Implementation Details}
We use two main histopathology image datasets in our experiments: GlaS~\cite{glas} of glands and MoNuSeg~\cite{monuseg} of nuclei. We uniformly crop patches of size $384\times 384$, with a fixed step size 64 along the width and height, from all the images. The GlaS dataset has 85 images (1697 patches) for training and 80 images (1559 patches) for test. We use 8 style clusters and 5 content clusters to study the data distributions. The MoNuSeg dataset has 30 high-resolution images, with 16 images (1296 patches) of 4 organs for training and 14 images (1134 patches) of 7 organs for test (note that 3 test organs are not seen in training). We use 7 style clusters and 3 content clusters. 
% We choose these cluster numbers since using more clusters will make some $H_{ij}$'s contain very few patches, unnecessarily increasing computation costs and potential statistic errors. 
In both the datasets, we treat either the training set or a subset of the training set as the labeled set and the test set or a subset of the training set (by ignoring the labels of this subset) as the unlabeled set in our experiments.

We use four segmentation models: DCN \cite{dcn}, MildNet \cite{graham2019mild}, FullNet \cite{Qu2019miccai}, and CIA-Net \cite{zhou2019cia}. Here, DCN and MildNet are common segmentation models. FullNet and CIA-Net attain state-of-the-art performance on the GlaS and MoNuSeg datasets, respectively. The default value of the probability $R_a$ is set as 0.15. The weights $w_1$, $w_2$, and $w_3$ in the loss function of Eq.~(\ref{eq:totalLoss}) are set as 0.002, 1, and 10, respectively. Basic augmentation operations (e.g., flipping, rotation) are applied in all the experiments (denoted as ${\rm Aug_{basic}}$).

\noindent
\subsection{Results}
Our experiments consist of three parts. In qualitative results, we show that our image generation module can effectively generate realistic images and cluster patches based on the style and content similarity. In quantitative results, we evaluate the effectiveness of our new method on improving segmentation performance under both the inductive learning and transductive learning settings of SSL scenarios. In ablation study, we evaluate the model sensitivity to different generation algorithms and policies.

\begin{figure}[!t]
    \centering
    \includegraphics[width=\columnwidth,height = 8.4cm]{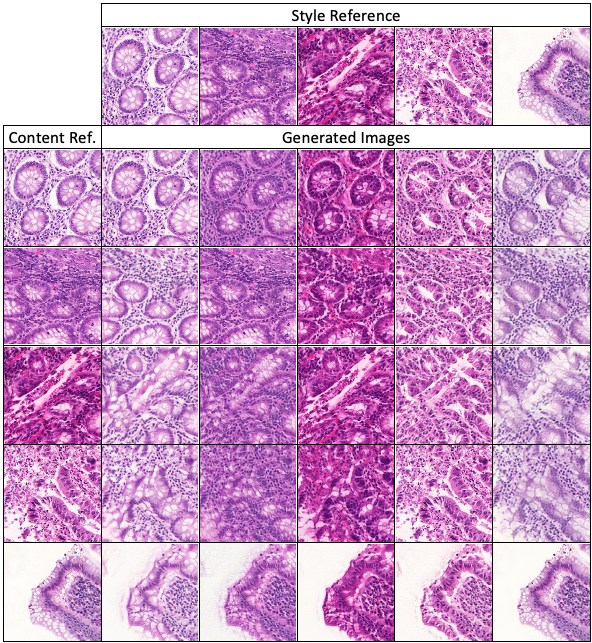}
    \caption{Examples of style transfer results for GlaS datasets. 
    }
    \label{fig:result1-1}
\end{figure}

\noindent{\bf Qualitative Results.}
Fig.~\ref{fig:result1-1} gives examples of generated images for the GlaS dataset. The patches in the same row are from the same content reference image patch (i.e., using the same $c_a$ given by the leftmost patch). The patches in the same column are generated with the same style. The results show our method can generate diversified realistic-looking images.
% Fig.~\ref{fig:supp2} shows differences of our model and MUNIT~\cite{munit} with an example. The MUNIT-generated image does not look realistic and the target style is not applied well (see the areas circled by green curves). In comparison, our method applies tissue styles better to the suitable positions.

% Fig.~\ref{fig:result1-2} shows the effectiveness of our Image Generation Module for style clustering. We use t-SNE~\cite{tsne} to visualize the style vector space. The style vectors are extracted from all the original images. The style vectors of benign glands and malignant glands do not have the same distributions, and are grouped into different areas. In a local area, the image patches tend to show similar appearance. In Fig.~\ref{fig:framework1}, one can see that in the same content cluster (e.g., in the same row $C_i$), the patches tend to have  similar shapes (e.g., round, irregular).
 
% \begin{figure}[]
%     \centering
%     \includegraphics[width=\columnwidth,height = 2.5cm]{supp2.pdf}
%     \caption{Comparison with MUNIT~\cite{munit} on single domain style transferring. }
%     \label{fig:supp2}
% \end{figure}

% \begin{figure}[t]
% \begin{center}
%   \includegraphics[width=1\linewidth,height = 4.cm]{fig7.png}
% \end{center}
%   \caption{Image patches in the GlaS dataset~\cite{glas} with similar appearance are locally grouped together, and different disease grades are in different areas of the style space.}
% \label{fig:result1-2}
% \end{figure}
\noindent{\bf Quantitative Results.}
In the inductive learning setting, we randomly choose 50\% of the patches from the original training set as labeled data $S_{label}$, with the rest of the training set ignoring their annotation as the unlabeled data $S_{unlabel}$, for extracting style information to help improve segmentation performance. Tables~\ref{tb1} and \ref{table:3} show the results. First, one can observe that with only 50\% labeled data, our SSL method can attain performance better than the results with full annotation for both the networks used. 
Second, comparing to the other known SSL methods (RA~\cite{RA} and CCT~\cite{cct}), our method yields the best segmentation performance with 50\% labeled data used.
Third, our method can be applied to any segmentation networks. In comparison, the SSL algorithm in CCT~\cite{cct} was designed upon a specific auto-encoder segmentation model structure.

Transductive learning means in the training process, the test images are shown as unlabeled images to the model (i.e., $S_{unlabel}$ is the original test set). In biomedical imaging, transductive learning has found wide applications. For example, in high-throughput experiments, a large amount of images (with potentially different styles) needs to be accurately segmented for further analysis. Transductive learning can be very useful in such scenarios. Tables~\ref{table:3} shows the segmentation results. We evaluate the segmentation models (DCN \cite{dcn}, MildNet \cite{graham2019mild}, and CIA-Net \cite{zhou2019cia}) with \emph{${\rm Aug_{basic}}$} on our proposed \emph{random generation} and \emph{mixed policy}. First, our method can effectively boost the performance of all the models, yielding better results than the state-of-the-art performance. Second, 
% further improvement can be made with our generation policy, indicating that this generation policy is as important as the data generating algorithm. Third, 
our method works well especially for the rare hard cases (e.g., unseen organs in MoNuSeg).

\begin{table}[t]
\caption{Inductive segmentation results of the GlaS dataset.}
\begin{center}
\resizebox{\columnwidth}{!}{
\begin{tabular}{|c|c|c|c|c|c|c|c|}
\hline
Labeled Data           & Setting                   & Model   &                            & F1-A  & F1-B  & Dice-A & Dice-B \\ \hline
\multirow{4}{*}{100\%} & \multirow{4}{*}{Baseline} & MildNet & w/o SSL                    & 0.914 & 0.844 & 0.913  & 0.836  \\ \cline{3-8} 
                       &                           & FullNet & w/o SSL                    & 0.924 & 0.853 & 0.914  & 0.856  \\ \cline{3-8} 
                       &                           & RA      & w/o SSL                    & 0.921 & 0.855 & 0.904  & 0.858  \\ \cline{3-8} 
                       &                           & CCT     & w/o SSL                    & 0.888 & 0.780 & 0.887  & 0.829  \\ \hline
\multirow{6}{*}{50\%}  & \multirow{4}{*}{Baseline} & MildNet & w/o SSL                    & 0.909 & 0.829 & 0.904  & 0.832  \\ \cline{3-8} 
                       &                           & FullNet & w/o SSL                    & 0.913 & 0.852 & 0.908  & 0.849  \\ \cline{3-8} 
                       &                           & RA      & SSL       & 0.916 & 0.862 & 0.897  & 0.856  \\ \cline{3-8} 
                       &                           & CCT     & SSL  & 0.847 & 0.793 & 0.845  & 0.821  \\ \cline{2-8} 
                       & \multirow{2}{*}{Ours}     & MildNet & SSL (Ours)                 & 0.917 & 0.840 & 0.905  & 0.845  \\ \cline{3-8} 
                       &                           & FullNet & SSL (Ours)                 & {\bf 0.925} & {\bf 0.862} & {\bf 0.919}  & {\bf 0.863}  \\ \hline
% \multirow{6}{*}{30\%}  & \multirow{4}{*}{Baseline} & MildNet & w/o SSL                    & 0.909 & 0.778 & 0.905  & 0.818  \\ \cline{3-8} 
%                       &                           & FullNet & w/o SSL                    & 0.909 & 0.835 & 0.894  & 0.839  \\ \cline{3-8} 
%                       &                           & RA      & SSL (Active Learning)      & 0.909 & 0.843 & 0.890  & {\bf 0.855}  \\ \cline{3-8} 
%                       &                           & CCT     & SSL (Consistency Learning) & 0.823 & 0.766 & 0.840  & 0.785  \\ \cline{2-8} 
%                       & \multirow{2}{*}{Ours}     & MildNet & SSL (Ours)                 & 0.913 & 0.822 & 0.902  & 0.837  \\ \cline{3-8} 
%                       &                           & FullNet & SSL (Ours)                 & {\bf 0.915} & {\bf 0.855} & {\bf 0.914}  & 0.853  \\ \hline
\end{tabular}
}
\end{center}
\label{tb1}

\end{table}

\begin{table}[t]
\caption{Transductive (top) and inductive (bottom) segmentation results of the MoNuSeg dataset.}
\begin{center}
\setlength{\tabcolsep}{3mm}
\resizebox{\columnwidth}{!}{
\begin{tabular}{|l|c|c|c|c|}
\hline
 & AJI  & AJI  & F1  & F1         \\ \hline
 & (Seen Organ) &  (Unseen Organ) &  (Seen Organ) &  (Unseen Organ)        \\ \hline
DCN  ${\rm Aug_{basic}}$      & 0.581            & 0.550           & 0.820             & 0.804 \\ \hline
DCN  Mix. Policy             & {\bf 0.594}            & {\bf 0.579}           & {\bf 0.829}             & {\bf 0.806} \\ \hline
MildNet  ${\rm Aug_{basic}}$    & 0.585            & 0.566           & 0.829             & 0.821 \\ \hline
MildNet  Mix. Policy         & {\bf 0.601}            & {\bf 0.594}           & {\bf 0.841}             & {\bf 0.833} \\ \hline
CIA-Net  ${\rm Aug_{basic}}$    & 0.613            & 0.631           & 0.824            & 0.846 \\ \hline
CIA-Net  Mix. Policy         & {\bf 0.632}            & {\bf 0.650}           & {\bf 0.857}             & {\bf 0.851} \\ \hline
\hline
\hline
CIA-Net  ${\rm Aug_{basic}}$ (50\%) & 0.591 & 0.603 & 0.818 & 0.830 \\ \hline
CIA-Net  Mix. Policy (50\%)                        & {\bf 0.625} & {\bf 0.615} & {\bf 0.850} & {\bf 0.831} \\ \hline
CIA-Net  ${\rm Aug_{basic}}$ (30\%)                  & 0.479 & 0.354 & 0.767 & 0.732 \\ \hline
CIA-Net  Mix. Policy (30\%)                        & {\bf 0.507} & {\bf 0.441} & {\bf 0.767} & {\bf 0.762} \\ \hline
\end{tabular}
 }

\end{center}

\label{table:3}
\end{table}

% \subsection{Ablation Study.}
\noindent {\bf Ablation Study.}
% \noindent {\it Generation Algorithms.}
Tables~\ref{otherstyle} shows the DCN model's performance on the GlaS dataset. The benefit to segmentation performance under the transductive setting of our Image Generation Module is compared to other known style transfer methods, including image-based (e.g., Gaytz et al.~\cite{gatys} and domain-based (e.g., MUNIT~\cite{munit}) methods. For fair comparison, we use our \emph{random generation} policy. One can see that our framework can achieve the best performance. 

% \noindent {\it Generation Policies.}
The contribution of each component in our generation policy is evaluated, including content matching (CM), distribution matching (DM), and hard case covering (HC). 
The performance is affected when only one policy is used or CM is not applied. The three components of our policy can help improve segmentation performance in a complementary manner. 
% Better performance can be achieved by applying our \emph{ensemble policy}.

% \noindent {\it Optimal Parameters.}
% Fig.~\ref{fig8} shows some key sensitive parameters of our method: the probability $R_a$ (top) and the weight for combining our two new policies (bottom). We evaluate the Dice scores of the DCN network on the GlaS test set with different values. In the top part of Fig.~\ref{fig8}, we show that the optimal $R_a$ value is 0.15. In a wide range of $R_a\in [0.05,0.5)$, our generation module can be beneficial to segmentation tasks. In the bottom part of Fig.~\ref{fig8}, we show that the optimal weight for combining our two generation policies (DM and HC) is 50\% --- 50\%, with which the best performance is achieved.

\begin{table}[]
 \caption{Ablation study for our Image Generation Module and generation policy.}
 \begin{center}

% \setlength{\tabcolsep}{0.5mm}
% \resizebox{\columnwidth*1/2}{!}{
\begin{tabular}{|l|c|c|c|c|}
\hline
     & F1-A  & F1-B  & Dice-A  & Dice-B \\ \hline
Baseline (${\rm Aug_{basic}}$)           & 0.923 & 0.822 & 0.910   & 0.826  \\ \hline
Gaytz et al.~\cite{gatys}               & 0.917 & 0.829 & 0.904   & 0.827  \\ \hline
% WCT~\cite{wct}                 & 0.922 & 0.816 & 0.916   & 0.823  \\ \hline
MUNIT~\cite{munit}               & 0.923 & 0.818 & 0.912   & 0.827  \\ \hline
Ours (Random Policy)                &  0.924 & 0.824 & {\bf 0.918}   &  0.839 \\ \hline \hline
Ours (CM + DM)          & 0.925 & 0.841 & 0.915   & 0.833  \\ \hline
Ours (CM + HC)          & 0.922 & 0.832 & {\bf 0.918}   & 0.836  \\ \hline
Ours (DM + HC)          & 0.925	 & 0.843  &	0.913 &	0.841   \\ \hline
Ours (Mix. Policy)        & {\bf 0.926} & {\bf 0.850} & 0.915   & {\bf 0.848}  \\ \hline
 \end{tabular}

 \label{otherstyle}

 \end{center}
\end{table}

% \begin{table}[]
% \begin{center}

%  \caption{Ablation study for our generation policies.}
% \capbtabbox{
% % \setlength{\tabcolsep}{.mm}
% % \resizebox{\columnwidth*1/2}{!}{
% \begin{tabular}{|l|c|c|c|c|}
% \hline
%                       & F1-A  & F1-B  & Dice-A  & Dice-B \\ \hline
% Mix. Policy        & {\bf 0.926} & {\bf 0.850} & 0.915   & {\bf 0.848}  \\ \hline
% CM + DM          & 0.925 & 0.841 & 0.915   & 0.833  \\ \hline
% CM + HC          & 0.922 & 0.832 & {\bf 0.918}   & 0.836  \\ \hline
% Mix. w/o CM          & 0.925	 & 0.843  &	0.913 &	0.841   \\ \hline \hline
% Ens. Policy       & 0.925 & 0.852 & 0.921   & 0.856  \\ \hline
%  \end{tabular}
%   \label{tb5}}
    
% \end{center}
% \end{table}

% \begin{figure}[]
%     \centering
%     \includegraphics[width=6cm,height = 3.5cm]{fig6_1.png}
%     \includegraphics[width=6cm,height = 3.cm]{fig6_2.png}
%     \caption{Ablation study for determining the optimal probability value $R_a$ and the weight for mixing our two policies. }
%     \label{fig8}
% \end{figure}

\section{Conclusions}
In this paper, we proposed a new unlabeled data guided semi-supervised learning framework for histopathology image segmentation. We designed (1) a style matching loss in our image generation module for image-based style transfer and exploring data distributions with concise style representations, and (2) new policies for guiding our image generation procedure. The effectiveness of our method was demonstrated by comprehensive experiments on two datasets. 

\vspace{0.15in}
\noindent{{\bf Acknowledgement.} This research was supported in part by NSF Grant CCF-1617735.}

% \bibitem{b1} G. Eason, B. Noble, and I. N. Sneddon, ``On certain integrals of Lipschitz-Hankel type involving products of Bessel functions,'' Phil. Trans. Roy. Soc. London, vol. A247, pp. 529--551, April 1955.
% \bibliography{bib}
% \end{thebibliography}
\bibliographystyle{IEEEtran}
% \bibliography{ref}

\end{document}